%% file: main.tex
\documentclass[10pt,journal,letterpaper]{IEEEtran}
\usepackage[T1]{fontenc}
\usepackage[utf8]{inputenc}
\usepackage{amsmath,amssymb}
\usepackage{graphicx}
\usepackage{booktabs,tabularx,array}
\usepackage{microtype}
\usepackage{placeins}
\usepackage{xurl}
\usepackage{cite}
\usepackage[dvipsnames]{xcolor}
\usepackage{soul}
\usepackage[hidelinks,unicode]{hyperref}
\renewcommand{\thesection}{\arabic{section}}

\hypersetup{pdftitle={Capturing LLM Utility While Managing LLM Fallibility}}
\title{Maximizing LLM Utility While Managing LLM Fallibility}

\author{Ekta U. Samani$^{1}$ and Christopher G. Atkeson$^{1}$
\thanks{This work was supported by the Manufacturing Futures Institute at Carnegie Mellon University.}
\thanks{$^{1}$E. U. Samani and C. G. Atkeson are with the Robotics Institute at Carnegie Mellon University,
        {\tt\small \{esamani,cga\}@andrew.cmu.edu}}%
}

\begin{document}
\raggedbottom
\maketitle
\begin{abstract}
\input{sections/abstract.tex}
\end{abstract}

\input{sections/01_introduction.tex}
\input{sections/02_ways.tex}
\input{sections/03_estimating.tex}

\input{sections/04_reliance.tex}
\input{sections/05_requirements.tex}
\input{sections/06_synthesis.tex}


\addtolength{\textheight}{-8cm}
\bibliographystyle{IEEEtran}
\bibliography{references}
\end{document}

%% file: sections/abstract.tex
Large language models (LLMs) can supply autonomous systems with plans, predictions, and judgments, but these contributions are sometimes wrong. This review asks how a system can maximize an LLM's utility while managing its fallibility. Our answer is to adapt the system's reliance on the LLM as evidence about the LLM's performance accumulates. Systems built from imperfect components, such as noisy sensors and faulty software, have long worked in this way, and the classical methods they use can be applied to LLM-based systems as well. A system can change its reliance on an LLM in several ways, depending on the LLM's role and on the other components in the system. Each change needs an estimate of the LLM's reliability, and the LLM's role and the kind of contribution it makes determine what to estimate. Classical methods from decision theory, estimation, and online learning then turn these estimates into reliance decisions. These decisions can reduce how often LLM errors affect the system. However, some errors matter even when they are rare. An independently verified mechanism can prevent such errors from being executed. Without one, adapting reliance using reliability estimates can still keep their risk within a bound. Studies on adapting reliance on LLMs are emerging but limited, and the reviewed comparisons suggest that adapting reliance can improve system performance.

%% file: sections/01_introduction.tex
\section{Introduction}
\label{sec:intro}

Large language models (LLMs) can supply plans~\cite{zeng2023socratic,singh2023progprompt,huang2022language}, designs, and judgments~\cite{liu2023reflect}, which we call LLM contributions, to robotic systems, and the gains from using them can be large. For example, COCO-TAMP \cite{kim2026cocotamp} uses LLM knowledge about where household objects are usually stored and which objects tend to be found together to guide a robot’s search. Combining this knowledge with Bayesian state estimation reduced average planning and execution time by 62.7\% in simulation compared with a baseline without this knowledge. However, a separate variant that asked the LLM to update location beliefs directly failed in three of five simulated trials with randomized object placements, while variants using Bayesian updates completed the tasks. These results illustrate both the value of LLM knowledge and the importance of how a system uses it. This review therefore asks how a system can maximize the utility of imperfect LLMs while managing their fallibility.

Systems built from imperfect components, such as noisy sensors and faulty software, faced the same question long before the advent of LLMs. Such systems tolerate these imperfections by limiting their \emph{reliance} on any one component. In fault-tolerant software, for example, a system can rely on the majority vote of multiple versions of a program, so that a faulty version is overruled \cite{knight1986experimental}. Methods from estimation, control, and online learning also help adapt reliance on imperfect components as evidence about each component's performance accumulates. For example, an adaptive Kalman filter \cite{mehra1970identification} weights each sensor by the uncertainty it estimates from that sensor's measurements. A contextual bandit \cite{li2010contextualbandit} learns which source works best in each context and a discounted bandit \cite{garivier2008upperconfidence} gives old outcomes less weight so that it can track a non-stationary source. LLM contributions can also be uncertain, context-dependent, and non-stationary \cite{ahn2022saycan, openai2026gpt56}. An LLM-based system can therefore use the same methods to adapt its reliance on an LLM. For example, the system can weight the LLM's contributions by their past performance, verify them before acting on them, or stop using the LLM altogether. Our hypothesis is that adapting reliance according to evidence about the LLM's reliability can improve system performance.

\begin{figure*}[!t]
\centering
\includegraphics[width=\textwidth]{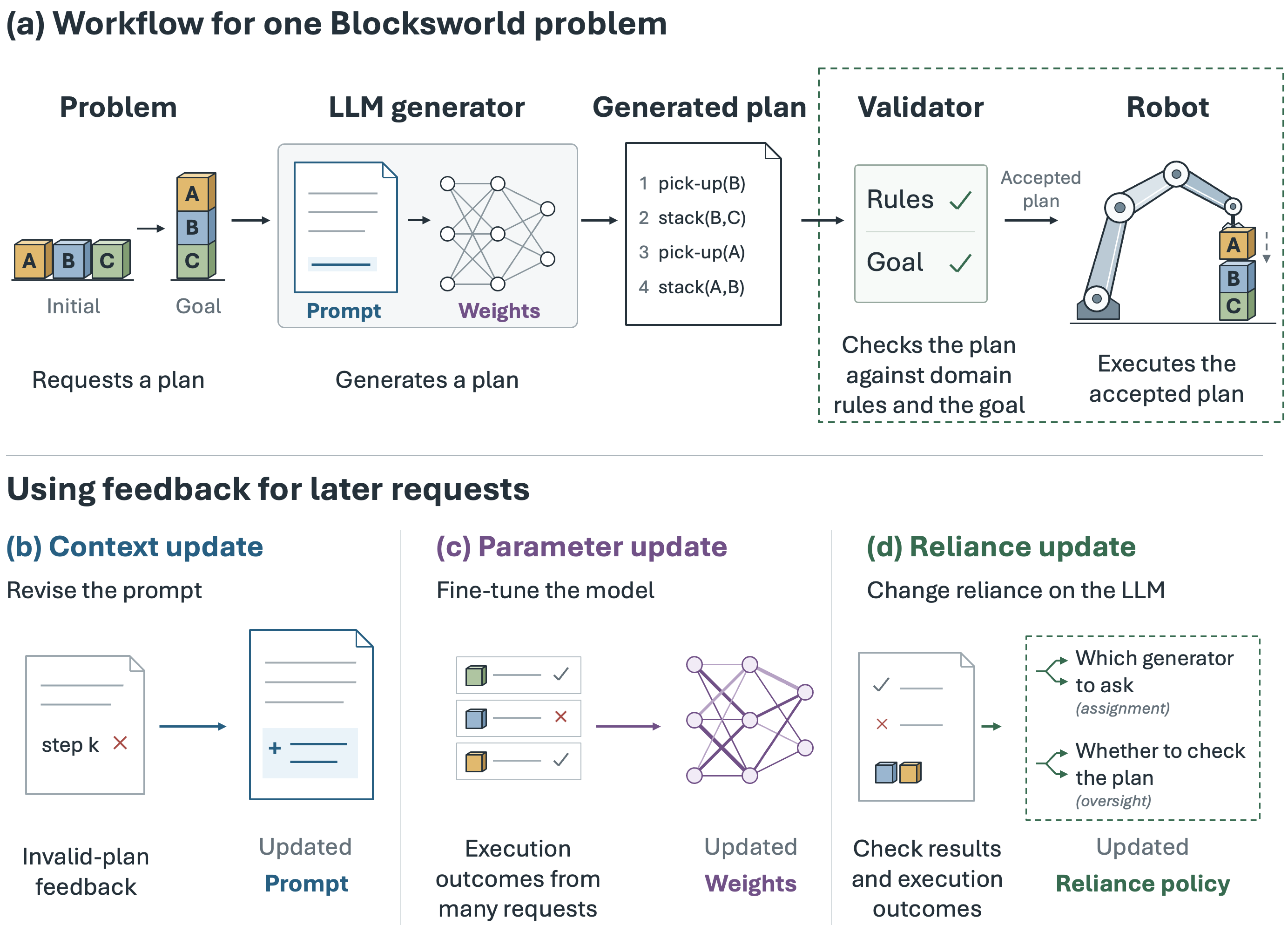}
\caption{Blocksworld example used throughout this review, adapted from \cite{valmeekam2023large}. For one Blocksworld problem, the LLM generates a plan and a validator checks it. The robot executes an accepted plan as shown in (a). Feedback generated from this can be used in three different ways: (b) adding invalid-plan feedback to the prompt, (c) fine-tuning the model on outcomes from many such problems, and (d) using validator results and execution outcomes as evidence to change how the surrounding system relies on the LLM.}
\label{fig:1}
\end{figure*}

Prior work already explores improving system performance by using the outcomes of LLM contributions to improve future LLM contributions. As shown in Fig.~\ref{fig:1}(b) and Fig.~\ref{fig:1}(c), feedback can be added to subsequent prompts \cite{huang2022inner,kobalczyk2026lilo}, or aggregated across many instances into fine-tuning data that updates the model's parameters \cite{boyle2025robotxr1}. Reinforcement learning from human feedback (RLHF) aggregates human feedback in the same way to improve the LLM \cite{ouyang2022training}. 

However, work on adapting reliance on LLM contributions that may be erroneous is emerging and limited \cite{cisse2025languagebased,chen2026labo,chen2026caliper,sridhar2026correlationaware, paudel2026object,kiyani2026trust}. Existing surveys address robotic applications of foundation models~\cite{firoozi2025foundation}, LLM uncertainty \cite{shorinwa2025survey}, planning \cite{wei2025plangenllms,tantakoun2025llms}, routing \cite{moslem2026dynamic}, and safeguards \cite{kojima2025comprehensive,dantas2026safe}. This review complements these surveys by showing how to apply classical methods to reliance decisions across LLM roles (Fig.~\ref{fig:1}(d)). Section~\ref{sec:ways} shows the ways in which we can change reliance on an LLM, depending on the role it plays in the system and on the other components in the system. Making such changes requires estimating the LLM's reliability from evidence (Section~\ref{sec:estimating}) and then turning those estimates into reliance decisions with classical methods (Section~\ref{sec:reliance}). Some LLM errors matter even when they are rare. Section~\ref{sec:requirements} shows what a system needs to prevent them from being executed and, when it cannot, how to keep their risk within a bound.

%% file: sections/02_ways.tex
\section{Ways to change reliance}
\label{sec:ways}

How a system might adapt its reliance on an imperfect LLM depends on the job the LLM does in the system and on what other components are present in the system. Consider the Blocksworld planning task in Figure~\ref{fig:1}, which we use as an example throughout this review. Valmeekam et al.~\cite{valmeekam2023large} ask GPT-4 to generate a sequence of actions that rearrange the blocks into a goal configuration. An external symbolic validator 
checks each plan against the task's domain rules and goal, returning feedback for revision when the plan is invalid. In another setup, the study also uses GPT-4 to judge a plan as valid or invalid, instead of the symbolic validator. The same GPT-4 acts as a \textit{planner} when it generates plans and as an \textit{evaluator} when it judges them. We call the job an LLM does in a system its functional \emph{role}. Across the applications we reviewed, LLMs contribute in the eight roles listed in Table~\ref{tab:roles}.

\input{tables/table_roles}

If we introduced another planner in the Blocksworld example, the system could adapt its reliance on the GPT-4 planner by changing \textit{assignment}: it could decide which of the two planners to use for a problem based on their performance. Similarly, the system could adapt its reliance on the GPT-4 evaluator by deciding when to use it or the external validator, based on the performance of the former.   If the external validator were expensive to run, the system could instead change \textit{oversight}: it could decide how often to check the GPT-4 planner's plans, based on their past validity. LABO, for example, decides when to run a costly real experiment to check its LLM's predictions~\cite{chen2026labo}. In some systems, the LLM's contribution assists a process that can otherwise run without it. For example, a system can stop asking a search shaper LLM that guides an optimizer~\cite{samani2026programming} for guidance if the LLM is unhelpful, because the optimizer can continue the search on its own. In such systems, adapting reliance can also be done by changing \textit{invocation}: deciding when to ask an LLM for a contribution based on how helpful its contributions have been so far. In each case, a system can decide to rely less on an imperfect LLM only when it can rely more on other components within the system. Table~\ref{tab:dimensions} lists six ways, or \textit{dimensions}, in which reliance can change, along with what other components the system must have to enable changes in those ways. 

\input{tables/table_dimensions}

Table~\ref{tab:mapping} shows how a system can change its reliance on an LLM in each functional role (from Table~\ref{tab:roles}) along each reliance dimension (from Table~\ref{tab:dimensions}). For example, if a system with an LLM controller has a fallback controller, it can change assignment by handing control to the fallback. If the system can check each command before execution, it can change oversight. If it can limit how far a command moves the robot, independently of the LLM, it can change latitude. Note that a system can also change its reliance along more than one of these dimensions at the same time~\cite{wang2026openloop}.

\input{tables/table_mapping}

%% file: tables/table_roles.tex
\begin{table}[!t]
\caption{Functional roles of LLM contributions}
\label{tab:roles}
\centering\footnotesize
\setlength{\tabcolsep}{3pt}
\begin{tabular}{@{}>{\raggedright\arraybackslash}p{50pt}>{\raggedright\arraybackslash}p{94pt}>{\raggedright\arraybackslash}p{96pt}@{}}
\toprule
\textbf{Role} & \textbf{What the LLM contributes} & \textbf{Example} \\
\midrule
Formulator / architect & An objective, reward, specification, heuristic, or code & An LLM writes reward functions for learning~\cite{ma2024eureka}, planning specifications~\cite{liu2023llmp}, and policy and planning code~\cite{liang2023code,huang2023voxposer}.\\
\addlinespace[2pt]
Candidate proposer & A candidate design, configuration, hypothesis, or solution & An LLM proposes batches of monomer sequences~\cite{reinhart2024large}. \\
\addlinespace[2pt]
Search shaper & A prior, constraint, ranking, region, or other guidance for a search & An LLM guides Bayesian optimization of 3D print settings~\cite{samani2026programming}. \\
\addlinespace[2pt]
Consequence modeler & A prediction of an outcome, property, feasibility, or uncertainty before it is measured & An LLM predicts the electrical conductivity of polymer composite films~\cite{chen2026labo}. \\
\addlinespace[2pt]
Evaluator & A score, judgment, critique, or check of a candidate, output, plan, or state & An LLM judges whether plans for rearranging blocks are valid~\cite{valmeekam2023large}. \\
\addlinespace[2pt]
Selector / planner & A choice of candidate, action, skill, tool, plan, or branch & An LLM generates plans for rearranging blocks~\cite{valmeekam2023large} or scores candidate actions for robot manipulation~\cite{ren2023robots}. \\
\addlinespace[2pt]
Coordinator & Routing, weighting, combination, or escalation among several contributors & An LLM combines top-ranked text responses from several LLMs into one response~\cite{jiang2023llmblender}. \\
\addlinespace[2pt]
Controller & A physical action or low-level control command & An LLM proposes velocity commands for a mobile robot~\cite{hafez2025safe}. \\
\bottomrule
\end{tabular}
\end{table}

%% file: tables/table_dimensions.tex
\begin{table}[!t]
\caption{Dimensions in which a system can change its reliance on an LLM}
\label{tab:dimensions}
\centering\footnotesize
\setlength{\tabcolsep}{3pt}
\begin{tabular}{@{}>{\raggedright\arraybackslash}p{46pt}>{\raggedright\arraybackslash}p{100pt}>{\raggedright\arraybackslash}p{94pt}@{}}
\toprule
\textbf{Dimension} & \textbf{Decision a system makes about using the LLM} & \textbf{What the system must have to change reliance along this dimension} \\
\midrule
Assignment & Whether the LLM or another source supplies the contribution & Another source for the needed contribution \\
\addlinespace[2pt]
Influence & How much weight the LLM's contribution gets & Other information to combine with the LLM contribution \\
\addlinespace[2pt]
Invocation & When to ask the LLM for a new contribution & A way to continue without a new LLM contribution \\
\addlinespace[2pt]
Oversight & Whether to check the LLM's contribution before using it & A way to check the LLM's contribution \\
\addlinespace[2pt]
Persistence & Whether an LLM contribution already in use stays in use & An LLM contribution in use and a way to continue without it \\
\addlinespace[2pt]
Latitude & Which options the LLM is allowed to choose from & A way to enforce permitted choices independently of the LLM \\
\bottomrule
\end{tabular}
\end{table}

%% file: tables/table_mapping.tex
\begin{table*}[!t]
\caption{How a system can change reliance on an LLM along each dimension in each functional role}
\label{tab:mapping}
\centering\footnotesize
\setlength{\tabcolsep}{3pt}
\begin{tabular}{@{}>{\raggedright\arraybackslash}p{58pt}>{\raggedright\arraybackslash}p{70.33pt}>{\raggedright\arraybackslash}p{70.33pt}>{\raggedright\arraybackslash}p{70.33pt}>{\raggedright\arraybackslash}p{70.33pt}>{\raggedright\arraybackslash}p{70.33pt}>{\raggedright\arraybackslash}p{70.33pt}@{}}
\toprule
\textbf{Functional role} & \textbf{Assignment} & \textbf{Influence} & \textbf{Invocation} & \textbf{Oversight} & \textbf{Persistence} & \textbf{Latitude} \\
\midrule
\textit{Formulator / architect} & Have a person or a library supply the formulation, or part of it & Weight LLM-written terms against hand-written ones, such as reward terms & Request a revised formulation after testing the current one & Test the formulation before adopting it, such as by running the code or training a policy~\cite{ma2024eureka} & Keep or replace an adopted formulation as test evidence comes in & Restrict changes to the formulation, such as by allowing edits only to the weights of fixed terms \\
\addlinespace[2pt]
\textit{Candidate proposer} & Have another generator supply candidates, such as through mutation or random sampling & Combine LLM-proposed candidates with others, such as by recombination, when their representations allow it & Request new candidates when other proposal operators stop improving the search & Screen candidates before use, such as in simulation or with a realizability checker~\cite{meng2025inprovf} & Keep candidates in the active search or discard them as test results arrive & Restrict the candidate space, such as through parameter bounds \\
\addlinespace[2pt]
\textit{Search shaper} & Obtain guidance from a heuristic or a person instead & Weight guidance against search evidence, such as measured objective values~\cite{chen2026caliper} & Request guidance when the search stalls, sooner after useful suggestions~\cite{cisse2025languagebased} & Test guidance before committing search budget, such as by sampling a suggested region & Keep or retire guidance as measured search gains change & Restrict guidance, such as by limiting the search space it may exclude \\
\addlinespace[2pt]
\textit{Consequence modeler} & Replace the prediction with other outcome information, such as a measurement & Combine predictions with other outcome information, such as surrogate estimates or measurements~\cite{chen2026labo} & Request a prediction when outcome uncertainty could change the next decision & Check predictions against observed outcomes, such as experimental measurements~\cite{chen2026labo} & Keep or retire predictions still guiding decisions as measurements arrive & Limit predictions to a permitted range, such as physically plausible values \\
\addlinespace[2pt]
\textit{Evaluator} & Have another judge, a validator, or a person supply the evaluation & Combine evaluations from several judges~\cite{zhao2026care} & Request an evaluation when cheaper checks leave a candidate unresolved & Spot-check evaluations against a reference, such as human labels & Keep or retire evaluations still guiding decisions as reference checks arrive & Restrict evaluation outputs, such as the permitted score range \\
\addlinespace[2pt]
\textit{Selector / planner} & Have another planner or a person make the choice or plan~\cite{ren2023robots} & Weight choices against task feasibility, such as skill affordance estimates~\cite{ahn2022saycan} & Request a new choice or plan when execution feedback warrants replanning & Check choices or plans before execution, such as with a symbolic validator~\cite{valmeekam2023large} & Keep executing a plan or stop partway, such as by committing to only its first steps & Restrict choices or plans, such as by limiting the available actions or skills \\
\addlinespace[2pt]
\textit{Coordinator} & Replace LLM coordination with fixed rules or a learned router & Blend coordination decisions from the LLM and a learned router, such as routing weights & Request new coordination decisions when the current allocation underperforms & Check coordination decisions against contributor capabilities or resource limits & Keep or retire coordination decisions as allocation results arrive & Restrict coordination decisions, such as which contributors may receive tasks or escalations \\
\addlinespace[2pt]
\textit{Controller} & Hand control to a fallback controller & Blend the LLM's command with a nominal one, as in shared control & Request new commands early when a verifier disagrees~\cite{wang2026openloop} & Check each command before execution, such as with a reachability check~\cite{hafez2025safe} & Stop executing the current commands when a verifier disagrees~\cite{wang2026openloop} & Restrict commands, such as by limiting how far they may move the robot \\
\bottomrule
\end{tabular}
\par\vspace{3pt}{\footnotesize\raggedright The system can change reliance on an individual contribution. It can also use accumulated evidence to change reliance on future contributions.\par}
\end{table*}

%% file: sections/03_estimating.tex
\section{Estimating reliability}
\label{sec:estimating}

The previous section showed which decisions a system can make to adapt its reliance on an imperfect LLM. This section shows how to estimate reliability of an LLM within a system to enable making those decisions.


Consider the Blocksworld example. Suppose the system must decide whether to use another planner instead of GPT-4, i.e., change its reliance along the \textit{assignment} dimension. It then asks: how often does GPT-4 produce a valid plan in a single shot? We call this question a reliability question. The quantity that answers this question, i.e., the valid-plan rate computed with the symbolic validator as the reference, is the \textit{reliability quantity}. In the study~\cite{valmeekam2023large}, GPT-4's single-shot plans were valid for 40 of 100 problems. Therefore, its reliability is estimated to be 0.40. For the GPT-4 evaluator, the reliability question has two parts: how often does it accept an invalid plan and how often does it mislabel valid plans as invalid? Therefore, the GPT-4 evaluator has two reliability quantities associated with it, the false-acceptance rate and the false-rejection rate. In the study, GPT-4 as a judge accepted 38 of 45 invalid plans (after the self-refinement pass) and rejected 1 of 55 valid ones.

The LLM's role narrows the general question of its reliability. For an evaluator, the reliability question is how often its verdicts agree with reference verdicts. For a consequence modeler, it is how close its predictions come to actual outcomes. For a controller, it is how often its actions lead to failure over repeated use. However, the role alone does not decide the reliability quantity to estimate. The quantity also depends on the contribution interface, i.e., the LLM's input and the form of its output. In the Blocksworld example, the GPT-4 evaluator returns a binary verdict, so its reliability quantities are the two error rates above. In G-Eval, by contrast, GPT-4 scores the summaries of other systems according to a written rubric, and its reliability quantity is rank correlation with human ratings~\cite{liu2023geval}. Similarly, more than one property can characterize a single contribution. A plan, for example, can be evaluated on validity, which a validator checks directly, or on task success, which shows only when the robot executes the plan. Once the role, interface, and property are known, an established method can often estimate the reliability quantity. Table~\ref{tab:quantities} gives a reliability question and quantity for several combinations of role, interface, and example property, and how to estimate the quantity. 
\input{tables/table_quantities}


Most estimates in Table~\ref{tab:quantities} need a record of the LLM's original contributions and a reference to score them against. When other components act on a contribution, the system's outcome reflects those components as well as the LLM~\cite{shah2023lmnav}. In the Blocksworld study, GPT-4 revises each plan the validator rejects. Its initial plans were valid for 40 of 100 problems, which estimates GPT-4's reliability as a planner. At the end of this refinement loop, 88 of 100 plans were valid, which estimates the loop's performance. Estimating the LLM's reliability therefore requires recording each contribution as the LLM produced it, before any other component acts on it.

An estimate of an LLM's reliability is only as good as its reference and the inputs on which we obtain it. We can use ground-truth labels as a reference directly, and a validator as a reference for the property it encodes. We can use a simulator as a reference once we have characterized its error against real outcomes for the property of interest~\cite{kennedy2001bayesian}. We can use a physical measurement as a reference if we account for its limits, since a measurement can be noisy, delayed, incomplete, or only a proxy for the property of interest~\cite{jadhav2025llm3d,jcgm2008evaluation}. We can also use these references to test many of the LLM's contributions before the LLM is deployed. Such tests estimate reliability only for inputs like those tested, so we should test on inputs that resemble those the system will meet in use.

%% file: tables/table_quantities.tex
\begin{table*}[!t]
\caption{Reliability questions and quantities by functional role, contribution interface, and example property, and how to estimate the quantities}
\label{tab:quantities}
\centering\footnotesize
\setlength{\tabcolsep}{4pt}
\begin{tabular}{@{}>{\raggedright\arraybackslash}p{112pt}>{\raggedright\arraybackslash}p{160pt}>{\raggedright\arraybackslash}p{228pt}@{}}
\toprule
\textbf{Functional role, contribution interface, and example property} & \textbf{Reliability question and quantity} & \textbf{How to estimate the quantity} \\
\midrule
\textit{Formulator / architect}\newline Specification\newline Example property: fit to the intended requirement & How well does the specification express the intended requirement?\newline Quantities: counts of omissions and detected contradictions, share of specification statements traced to requirement sources & Trace each specification statement to a requirement source. Then check which requirements the specification, under the domain assumptions, omits or contradicts, as in requirements engineering~\cite{zave1997four}. \\
\addlinespace[2pt]
\textit{Candidate proposer}\newline Candidate set\newline Example property: quality and variety of the set & How many good candidates does the set contain, and how novel and diverse are they?\newline Quantities: yield, quality, novelty, diversity, coverage under a common evaluator and budget & Score all candidates, from the LLM and other sources, with the same evaluator and budget. Measure novelty and coverage against a reference set, using ideation or quality-diversity metrics~\cite{shah2003metrics,pugh2016quality}. \\
\addlinespace[2pt]
\textit{Search shaper}\newline Suggested points or regions\newline Example property: improvement on the best value found so far & How often do suggested points, or points sampled from a suggested region, improve on the best value found so far?\newline Quantity: fraction of these points that improve on that value & Evaluate the objective at each suggested point, or at points sampled uniformly from each suggested region. Count those that improve on the best value. \\
\addlinespace[2pt]
\textit{Consequence modeler}\newline Dynamic model\newline Example property: prediction accuracy in the operating region & How accurately does the model predict new input--output data in the operating region?\newline Quantities: one-step and multistep prediction error & Compare the model's predictions with observed outcomes on new input--output data, at the horizon at which the model will be used, as in system identification~\cite{ljung2010perspectives}. \\
\addlinespace[2pt]
\textit{Consequence modeler}\newline Simulator or surrogate\newline Example property: agreement with observations & How does the model compare with observations?\newline Quantities: prediction error, model discrepancy & Pair the model's predictions with observations of the same inputs, and model the discrepancy between them~\cite{kennedy2001bayesian}. Account for observation error, and state the assumptions used to separate it from the discrepancy. \\
\addlinespace[2pt]
\textit{Consequence modeler}\newline Probability forecast\newline Example property: accuracy and concentration of forecasts & How well calibrated are the probabilities, and how concentrated and accurate are the forecasts?\newline Quantities: calibration, sharpness, proper scores, which answer different questions~\cite{gneiting2007strictly,gneiting2007probabilistic} & Record each forecast before its outcome resolves. Measure sharpness from the forecasts alone. Once outcomes resolve, check calibration and score each forecast with a proper scoring rule~\cite{dawid1984prequential}. \\
\addlinespace[2pt]
\textit{Evaluator}\newline Categorical verdict, such as valid or invalid plan, success or failure\newline Example property: correctness & How well do the verdicts separate the classes?\newline Quantities: false-acceptance rate among items that should be rejected, and false-rejection rate among items that should be accepted & Pair each verdict with a reference label, and compute both rates from the pairs. \\
\addlinespace[2pt]
\textit{Evaluator}\newline Ranking or score\newline Example property: correct ordering of items & How well do rankings agree with a reference, and how strongly are scores associated with human ratings?\newline Quantities: rank agreement, rank correlation with human ratings & Compare the LLM's rankings or scores with reference rankings or human ratings of the same items, using rank agreement or rank correlation~\cite{kendall1938new}. \\
\addlinespace[2pt]
\textit{Evaluator}\newline Numerical estimate of a physical quantity\newline Example property: accuracy & How does the estimate compare with a reference measurement of the same quantity?\newline Quantities: bias, repeatability, uncertainty & Compare repeated estimates of the same quantity with a reference measurement, and account for the reference's own uncertainty~\cite{jcgm2008evaluation}. \\
\addlinespace[2pt]
\textit{Selector / planner}\newline Plan\newline Example property: validity & How often do plans follow the domain rules and reach the goal?\newline Quantity: valid-plan rate & Check each original plan with a validator that encodes the intended property, such as VAL~\cite{howey2004val}. Report the fraction that pass. \\
\addlinespace[2pt]
\textit{Coordinator}\newline Routing or escalation decision\newline Example property: correct choice of contributor & How often does the LLM send an input to a contributor with the best result?\newline Quantity: fraction of inputs sent to such a contributor & Run every contributor on the same test inputs. Count the inputs on which the LLM chose a contributor with the best result. \\
\addlinespace[2pt]
\textit{Controller}\newline Command or command sequence\newline Example property: failure-free execution & How often does a command or sequence lead to a defined failure?\newline Quantity: failure probability per command or sequence & Record each command or sequence before any filter acts. Determine whether each original command or sequence would lead to the defined failure. Use physical trials without the filter where that is safe, or a simulation validated for the operating conditions. \\
\bottomrule
\end{tabular}
\end{table*}

%% file: sections/04_reliance.tex
\section{Adapting reliance using reliability estimates}
\label{sec:reliance}

The previous section showed how to estimate an LLM's reliability. This section shows how a system can use these estimates to change its reliance on the LLM.

Consider the Blocksworld example that uses GPT-4 as a planner and as a plan evaluator. Suppose the system wants to decide whether to check GPT-4's plans before execution, i.e., change its reliance along the \textit{oversight} dimension. Checking a plan would add the cost of the check, but it can prevent the robot from executing an invalid plan. A check performed by the GPT-4 evaluator can also be wrong, i.e., it might accept an invalid plan or reject a valid one. In the study, the GPT-4 evaluator checked 100 plans of which 45 were invalid plans. It accepted 38 of those 45 invalid plans and it rejected one of the 55 valid plans. Therefore, checking with the GPT-4 evaluator would prevent only seven invalid executions, at the cost of 100 checks and one wrongly rejected valid plan. Adapting reliance on the GPT-4 planner using the GPT-4 evaluator check requires weighing the planner's errors against the cost and errors of the evaluator: checking pays off only if the benefit of avoiding seven invalid plans is higher than the cost of checking 100 plans and rejecting one valid plan. In other words, the reliability estimates for the GPT-4 planner and evaluator can help change the reliance on the GPT-4 planner along the oversight dimension. More generally, classical methods from decision theory, estimation, and online learning tell us how to make the reliance changes shown in Table~\ref{tab:mapping}. We summarize these methods for each reliance dimension in Table~\ref{tab:methods}.

\input{tables/table_methods}

Studies so far show that adapting reliance on LLM contributions can improve system performance. For example, in BORA, a Bayesian optimizer searches on its own and asks its LLM for points to evaluate only when the search stalls~\cite{cisse2025languagebased}. The optimizer asks sooner when the LLM's earlier points performed well and later when they performed poorly. BORA found best values up to 67\% higher than an optimizer that relied on the LLM alone. Similarly, CALIPER decides during a multi-objective Bayesian optimization run whether to use LLM priors, by testing them against the measured objectives~\cite{chen2026caliper}. Where the LLM priors helped, CALIPER matched the performance of always using them. Where they hurt, however, always using them performed worse than CALIPER, which abstained from them. This study shows that LLM performance can vary across datasets. Adapting reliance on the LLM to its performance within the system therefore fares better than deciding in advance to always use its priors. In robotic object search, Paudel et al.~\cite{paudel2026object} select among fixed search strategies, including combinations of LLMs and prompts and a non-LLM alternative, using observed costs and offline replay after each trial. Their replay-assisted selection reduced search cost in simulation relative to ordinary UCB, which also adapts its selection. Kiyani et al.~\cite{kiyani2026trust} adapt oversight of an LLM evaluator by updating decision thresholds from feedback supplied by a stronger verifier. In Sudoku, this reduced how often a program checked proposed moves while retaining accuracy close to checking every move.

Every method in Table~\ref{tab:methods} helps a system decide how to change reliance on LLMs using estimates of the LLM's reliability, so its decisions are only as good as those estimates. To keep the estimates accurate, we need to account for four things. First, a system observes outcomes only for the LLM contributions it uses or checks~\cite{lakkaraju2017selective,auer2002nonstochastic,alon2015online}. In the Blocksworld example, a system that stops checking GPT-4's plans once they prove mostly valid would not notice if they later got worse. Likewise, a system that rarely uses an LLM with a low reliability estimate would not notice if that LLM improved. The same gap arises whenever a system's own decisions determine which outcomes it observes (e.g., when it skips a check, stops a contribution, or excludes a choice). We should therefore continue to collect evidence on the LLM's contributions and on every alternative source the system may rely on instead of the LLM. We can route a fraction of inputs to each alternative, check a random sample of contributions, or evaluate a second LLM on the same inputs. If the system routes inputs at random, it knows the probability with which each input reaches each alternative. Propensity weighting then weights each observed outcome by the inverse of that probability, so that the reliability estimates represent all inputs~\cite{swaminathan2015counterfactual}. For contributions that the system stops or choices that it excludes, evidence about their outcomes can come from tests before deployment or from validated simulation.

Second, a reliability estimate holds only for the LLM, the machine, and the kinds of inputs that produced its evidence. BORA, for example, asks its LLM for points only when its search stalls, so its evidence about the LLM comes from stalled searches~\cite{cisse2025languagebased}. We should therefore collect evidence on the inputs, and at the stages of a task, where the system will make each reliance decision. An LLM's reliability can also shift after a model update or a move to a new machine. Testing for such a shift is hard because other things often change at the same time. When a system whose LLM agents control 3D printing was moved to a new printer, for example, the documentation the agents received also changed~\cite{jadhav2025llm3d}. We should therefore record what changed before reusing earlier evidence, and estimate the LLM's reliability again after each change. When a shift may come without a known change, a discounted bandit~\cite{garivier2008upperconfidence} tracks such a shift by giving older outcomes less weight.

Third, a system may combine contributions from several LLMs, or use one LLM to check contributions from another LLM. The LLMs can make similar errors when they share a model, training data, or context. In one molecule-optimization study, for example, every tested LLM underestimated the same objective, so reweighting their predictions could not remove the error~\cite{chen2026caliper}. Reliability estimates built from the LLMs' agreement or checks therefore overstate their reliability. Before relying on that agreement or those checks, we should estimate each LLM's errors against a reference on common inputs. We can also model how their errors depend on one another, as CARE does when it combines LLM judges' scores~\cite{zhao2026care}.

Finally, an LLM's errors can be large and rare, and an estimate of its average error hides them. A system may fuse an LLM's predictions with measurements using a Kalman filter. The filter assumes normally distributed errors, so an occasional much larger error receives the same weight as an ordinary one~\cite{masreliez1977robust}. Therefore, we should use robust filters that model such errors and give them less weight~\cite{masreliez1977robust,ting2007kalman}. 

%% file: tables/table_methods.tex
\begin{table*}[!t]
\caption{Classical methods for deciding how to change reliance along each dimension, and how to apply each to an LLM}
\label{tab:methods}
\centering\footnotesize
\setlength{\tabcolsep}{3pt}
\begin{tabular}{@{}>{\raggedright\arraybackslash}p{42pt}>{\raggedright\arraybackslash}p{112pt}>{\raggedright\arraybackslash}p{254pt}>{\raggedright\arraybackslash}p{90pt}@{}}
\toprule
\textbf{Dimension} & \textbf{Classical method} & \textbf{How to apply it to an LLM} & \textbf{Related LLM system} \\
\midrule
Assignment & Algorithm selection assigns each input to the source predicted to perform best, given the input's features~\cite{rice1976algorithm}. & Test every source on inputs that resemble those met in use. Estimate each source's reliability on each type of input with the same criterion, such as its valid-plan rate per problem type. Send each new input to the source with the highest estimate for its type. & RouteLLM learns a router from offline preferences~\cite{ong2025routellm}. \\
\addlinespace[2pt]
 & A contextual bandit learns which source works best in each context~\cite{li2010contextualbandit,auer2002nonstochastic}. & Record each input's context, the chosen source, and its reward, such as task success or plan validity. Update each source's expected reward for that context, including rewards that arrive only after a costly check, such as a physical experiment. Choose the source for each new input with the bandit, which keeps trying alternatives. & CABS-D routes each query to one LLM, using selected rewards and surrogate information~\cite{sridhar2026correlationaware}. \\
\addlinespace[2pt]
 & A discounted bandit learns which source works best now by giving older outcomes less weight~\cite{garivier2008upperconfidence}. & Record each chosen source's reward. If an LLM's reliability may shift, such as after a model update or a move to a new machine, give older rewards less weight. Keep sending some inputs to every source. Route each input by the current estimates. & ParetoBandit routes each query to one LLM and gives older rewards less weight, so it can track shifts in LLM quality and price. It was tested offline, on precomputed rewards with injected shifts \cite{tabernermiller2026paretobandit}. \\
\addlinespace[2pt]
 & Adjustable autonomy assigns each decision to the automation or to a person, counting the cost of transferring control~\cite{scerri2002adjustable,parasuraman2000model}. & Calibrate the LLM's confidence in its choice on inputs that resemble those met in use. Also estimate the person's decision quality. Hand the decision to the person when their expected improvement over the LLM's choice outweighs the cost and delay of transferring control. & KnowNo asks the user whenever more than one action qualifies~\cite{ren2023robots}. \\
\midrule
Influence & Weighted Majority and Hedge weight each source's vote by its accumulated loss~\cite{littlestone1994weighted,freund1997decisiontheoretic}. & Record each source's error on the same inputs, such as each judge's error rates. Give less weight to sources with higher accumulated loss. Check these inputs for errors shared by every source. If the system executes the combined choice, act on a single source's choice for some inputs to observe each source's own loss. & CALIPER sets each LLM expert's weight from its discounted errors on measured objectives~\cite{chen2026caliper}. \\
\addlinespace[2pt]
 & Kalman or Bayesian fusion weights the sources by the estimated uncertainty and correlation of their errors~\cite{kalman1960new}. & Estimate the bias and uncertainty of the LLM's predictions against measurements. Model the correlation between its errors and those of the other sources. Weight each source by the resulting error model. Update the error model as new measurements arrive, and reweight the sources. If the LLM sometimes errs by much more than usual, give such errors less weight with a robust filter~\cite{masreliez1977robust}. & LABO models a measured outcome as a scaled LLM prediction plus a discrepancy~\cite{chen2026labo}. \\
\midrule
Invocation & Value of information compares deciding without a new contribution with asking for one first~\cite{howard1966information,berger1985statistical}. & Record each requested contribution's outcome, such as whether a suggested point improved on the best value. Estimate the LLM's reliability from these outcomes at each point where the system may decide to ask, such as when its search stalls. Ask when the predicted gain in later decisions outweighs the cost of asking. & BORA asks its LLM when the search stalls, sooner after its points improved the best value. The rule is simpler than value of information~\cite{cisse2025languagebased}. \\
\midrule
Oversight & Value of information compares acting on the contribution with checking it first~\cite{howard1966information,berger1985statistical}. & Estimate the source's error rate and the check's error rates on that source's contributions, such as a judge's false-acceptance rate on a generator's plans. Check a contribution before using it when the expected cost of acting directly exceeds that of checking, counting the check's own errors. Keep checking some contributions even when their error rate is low. & LABO runs a real experiment when discrepancy dominates its uncertainty. The rule is simpler than value of information~\cite{chen2026labo}. \\
\midrule
Persistence & Sequential decision analysis compares continuing with switching or stopping~\cite{berger1985statistical}. & Estimate the chance that a contribution in use still succeeds, given its progress so far. Base this estimate on completed runs, such as tests before deployment or validated simulation. Continue, switch, or stop, whichever has the highest expected value after counting the costs of failure and of switching. & SV-VLA stops the current command chunk when its verifier detects disagreement during execution~\cite{wang2026openloop}. \\
\midrule
Latitude & Trust-region methods widen or narrow the region in which a model may propose steps. Classical versions compare predicted with actual improvement~\cite{conn2000trustregion}. Categorical versions measure the region by Hamming distance and resize it after successive successes or failures~\cite{wan2021think}. & If the LLM chooses within a region of permitted choices scored by a common objective, compare each choice with the best value so far. Widen the region after successive improvements, and narrow it after successive failures. If choices are excluded to protect safety, get evidence about them from outside operation, such as validated simulation or offline tests. & HOLLM selects regions from observed objective values and asks its LLM for points inside them. It does not resize regions after the LLM's successes or failures~\cite{schwanke2026improving}. \\
\bottomrule
\end{tabular}
\end{table*}

%% file: sections/05_requirements.tex
\section{Meeting critical requirements}
\label{sec:requirements}

The previous section provides methods to adapt reliance on fallible LLMs, but some LLM errors matter even when they are rare. This section asks how a system can keep such errors from reaching execution or keep their risk within a bound.

\subsection{With an independently verified mechanism}

In the Blocksworld example, the symbolic validator approves a plan only if the plan follows the domain rules and reaches the goal \cite{howey2004val}. The robot therefore executes only valid plans, regardless of how often GPT-4 provides invalid plans. Such an independently verified mechanism keeps invalid plans from execution without relying on the LLM's reliability, because it acts on whatever the LLM produces \cite{kambhampati2024llms}. Mechanisms of this kind can act before an LLM contribution is used, when it is executed, or while execution continues. A validator checks a contribution, such as a plan or code, before use and must never approve one that violates the requirement. A filter, such as a shield \cite{alshiekh2018safe}, a control barrier function \cite{ames2016control}, or a reachability check \cite{hafez2025safe}, acts at execution and changes the resulting action instead of rejecting the entire contribution. A filter blocks only the violations its model can predict, so the model must cover what the system can actually do. A runtime monitor acts during execution and hands control to a fallback when the system nears a violation \cite{hobbs2022run}. The monitor must switch early enough for the fallback to prevent the violation. Such a mechanism can stay in place while the system uses reliability estimates to change reliance on LLMs.

\subsection{Without an independently verified mechanism}

Many requirements that an LLM's contribution might violate lack an independently verified mechanism, or have one that is too expensive to run. A system may then check the LLM's contributions with a simulator or with another LLM, but both checks are fallible. A simulator's test of a plan's physical outcome does not guarantee anything until we characterize the simulator's error against real outcomes~\cite{kennedy2001bayesian}. A system can still bound the risk of violating a requirement by adapting its reliance on the LLM using reliability estimates. KnowNo~\cite{ren2023robots} applies this approach to an LLM-based robot planner. Given a user's instruction, the LLM proposes candidate next actions and scores each for how likely it is to be correct. When instructions are ambiguous, the robot has no way to verify which one the user means. The authors therefore calibrate a score threshold on example instructions whose correct actions are known, for a success probability that the user chooses. For each new instruction, the planner retains the candidate actions whose LLM scores exceed this threshold. Under the conformal prediction assumptions, this set contains the correct action with the specified probability. If only one action qualifies, the robot executes it. Otherwise, the robot asks the user to choose. The calibrated threshold thus determines when the robot relies on the LLM and when it seeks help, bounding the risk of task failure without verifying individual actions.

%% file: sections/06_synthesis.tex
\section{Synthesis}
\label{sec:synthesis}

This review shows how classical methods long used to manage imperfect components can help a system benefit from an LLM while managing its fallibility. Applying these methods requires connecting what the LLM contributes, how its reliability can be estimated, and how the system can change its reliance on it. The review makes these connections across LLM roles and explains what each reliance decision requires. Early comparisons show that decisions made this way can improve system performance~\cite{cisse2025languagebased,chen2026labo,chen2026caliper,sridhar2026correlationaware,paudel2026object,kiyani2026trust}. Adapting reliance can also reduce how often LLM errors affect the system. When errors that matter even when rare cannot be prevented, it can still keep their risk within a bound. With these methods, a system can adapt its reliance on an LLM in any of the roles in Table~\ref{tab:roles}, from proposing candidates to controlling a robot. Tables~\ref{tab:mapping}--\ref{tab:methods} lay out how to do so in each role, giving researchers a starting point for building such systems. Table~\ref{tab:mapping} shows how reliance on an LLM can change in each role, Table~\ref{tab:quantities} shows how to estimate the LLM's reliability depending on how it is used in the system, and Table~\ref{tab:methods} shows which classical method can be used to turn those estimates into a reliance decision. The methods considered in this review emphasize adapting reliance based on estimates of the LLM's past performance. An open question is how a system can combine the LLM's observed performance with domain knowledge about its contributions to decide when to increase or reduce reliance on it.